\documentclass[conference, a4paper]{ieeeconf}  
\usepackage[left=1.91cm,right=1.31cm,top=3.7cm,bottom=1.91cm]{geometry}
\pdfoutput=1

\IEEEoverridecommandlockouts                              

\usepackage{graphicx} 
\usepackage{epstopdf}

\usepackage{amsmath} 
\usepackage{amsthm}  
\usepackage{amsfonts}
\usepackage[dvipsnames]{xcolor}
\usepackage{tikz}
\usepackage{pgfplots}
\usetikzlibrary{shapes,arrows}
\usetikzlibrary{backgrounds}
\usepackage{multirow}
\usepackage[keeplastbox]{flushend}
\usepackage[utf8]{inputenc}    
\usepackage[T1]{fontenc}       
\usepackage[USenglish]{babel}             
\usepackage{silence}  							
\usepackage{csquotes}

\usepackage{url}
\usepackage{letltxmacro}
\LetLtxMacro{\autocite}{\cite}
\LetLtxMacro{\textcite}{\cite}

\usepackage{pgfplots}
\pgfplotsset{compat=1.9}  
\usepackage[capitalize]{cleveref}
\Crefname{figure}{Fig.}{Figs.}
\crefname{equation}{}{}
\Crefname{equation}{Equation}{Equations}

\usepackage{subcaption}
\usepackage{xstring}
\usepackage{xparse}
\usepackage{siunitx}

\theoremstyle{plain}
\newtheorem{definition}{Definition}
\newtheorem*{problem}{Problem}
\theoremstyle{remark}\newtheorem{remarkenv}{Remark}        
\newenvironment{remark}{\begin{remarkenv}}%
	{\hfill$\lozenge$\end{remarkenv}}            

\usepackage{booktabs}
\usepackage{tabularx}
\newcolumntype{Y}{>{\raggedright\arraybackslash}X}
\newcommand{\otoprule}{\midrule[\heavyrulewidth]}

\graphicspath{ {figures/} }

\title{\LARGE \bf
	Real-World Scenario Mining for the Assessment of Automated Vehicles
}

\author{Erwin de Gelder$^{1,2*}$, Jeroen Manders$^{1}$, Corrado Grappiolo$^{3}$, Jan-Pieter Paardekooper$^{1,4}$, \\ Olaf Op den Camp$^{1}$, Bart De Schutter$^{2}$%
\thanks{$^{1}$TNO, Integrated Vehicle Safety, Helmond, The Netherlands}%
\thanks{$^{2}$Delft University of Technology, Delft Center for Systems and Control, Delft, The Netherlands}%
\thanks{$^{3}$TNO, Data Science, The Hague, The Netherlands}%
\thanks{$^{4}$Radboud University, Donders Institute for Brain, Cognition and Behaviour, Nijmegen, The Netherlands}%
\thanks{$^{*}$Corresponding author. \newline E-mail address: {\tt\small erwin.degelder@tno.nl}}%
}

\newlength\figurewidth
\newlength\figureheight
\definecolor{egocolor}{RGB}{220, 200, 200}
\definecolor{othervehicle}{RGB}{200, 220, 200}
\definecolor{staticenvironment}{RGB}{200, 200, 220}
\newlength{\tagheight}
\newlength{\subjectwidth}
\newlength{\descriptionwidth}
\newlength{\tagtotalwidth}
\newlength{\egovehiclecruising}
\newlength{\egovehicleaccelerating}
\newlength{\egovehiclefollowing}
\newlength{\egovehiclechanging}
\newlength{\itempos}
\newlength{\timepos}
\newlength{\itemwidth}

\newlength{\tagsep}
\newlength{\otherfollowing}
\newlength{\otherchanging}
\newlength{\othernolead}
\newlength{\cutinheight}
\newlength{\arrowheight}

\tikzstyle{block}=[minimum height=\tagheight, align=center, anchor=north east]
\tikzstyle{tag}=[block, draw, anchor=north west]
\tikzstyle{tagitem}=[tag, minimum width=\itemwidth, text width=\itemwidth-1em]
\tikzstyle{tagitemtwo}=[tag, minimum width=2\itemwidth, text width=2\itemwidth-1em]
\tikzstyle{tagitemthree}=[tag, minimum width=3\itemwidth, text width=3\itemwidth-1em]
\tikzstyle{tagitemfour}=[tag, minimum width=4\itemwidth, text width=4\itemwidth-1em]
\tikzstyle{timeline}=[ultra thick, color=black]
\tikzstyle{showitem}=[dashed]
\tikzstyle{cutinline}=[ultra thick, color=red]

\usetikzlibrary{arrows,positioning}
\usetikzlibrary{arrows.meta}

\newcommand{\symboldec}[2]{#1^{-}\left(#2\right)}
\newcommand{\symbolinc}[2]{#1^{+}\left(#2\right)}
\newcommand{\symboldectarget}[3]{#1^{-}_{#3}\left(#2\right)}
\newcommand{\symbolinctarget}[3]{#1^{+}_{#3}\left(#2\right)}

\newcommand{\accelerationsymbol}{a}
\newcommand{\accelerationcruise}{\accelerationsymbol_\mathrm{cruise}}
\newcommand{\defsym}{\equiv}
\newcommand{\factorgoalmax}{\alpha_1}
\newcommand{\factorgoalmin}{\alpha_2}
\newcommand{\fonescore}{\mathrm{F1}}
\newcommand{\indextarget}{i}
\newcommand{\indextargetother}{j}
\newcommand{\lanechangethreshold}{\Delta_{\mathrm{l}}}
\newcommand{\lanechangespeed}{v_{\mathrm{lat}}}
\newcommand{\lineleftsymbol}{l}
\newcommand{\lineleft}[1]{\lineleftsymbol\left(#1\right)}
\newcommand{\lineleftdec}[1]{\symboldec{\lineleftsymbol}{#1}}
\newcommand{\lineleftinc}[1]{\symbolinc{\lineleftsymbol}{#1}}
\newcommand{\linelefttarget}[2]{\lineleftsymbol_{#2}\left(#1\right)}
\newcommand{\linelefttargetdec}[2]{\symboldectarget{\lineleftsymbol}{#1}{#2}}
\newcommand{\linelefttargetinc}[2]{\symbolinctarget{\lineleftsymbol}{#1}{#2}}
\newcommand{\linerightsymbol}{r}
\newcommand{\lineright}[1]{\linerightsymbol\left(#1\right)}
\newcommand{\linerightdec}[1]{\symboldec{\linerightsymbol}{#1}}
\newcommand{\linerightinc}[1]{\symbolinc{\linerightsymbol}{#1}}
\newcommand{\linerighttargetdec}[2]{\symboldectarget{\linerightsymbol}{#1}{#2}}
\newcommand{\linerighttargetinc}[2]{\symbolinctarget{\linerightsymbol}{#1}{#2}}
\newcommand{\linerighttarget}[2]{\linerightsymbol_{#2}\left(#1\right)}
\newcommand{\lontarget}[2]{x_{#2}\left(#1\right)}

\newcommand{\precision}{\mathrm{Precision}}
\newcommand{\recall}{\mathrm{Recall}}
\newcommand{\sample}{k}
\newcommand{\sampledummy}{\tau}

\newcommand{\sampleendinc}[1]{\sample_{\mathrm{end}}\left(#1\right)}
\newcommand{\samplehorizon}{\sample_{\mathrm{h}}}
\newcommand{\sampleinit}{\sample_{0}}

\newcommand{\samplescruising}{\sample_\mathrm{cruise}}
\newcommand{\speedsymbol}{v}
\newcommand{\speed}[1]{\speedsymbol\left(#1\right)}
\newcommand{\speedmax}[1]{\speedsymbol_\mathrm{max}\left(#1\right)}
\newcommand{\speedmin}[1]{\speedsymbol_\mathrm{min}\left(#1\right)}
\newcommand{\speeddec}[1]{\symboldec{\speedsymbol}{#1}}
\newcommand{\speedinc}[1]{\symbolinc{\speedsymbol}{#1}}
\newcommand{\speeddiff}{\Delta_{\speedsymbol}}
\newcommand{\speedtargetiabs}[2]{\speedsymbol_{#2}^{\mathrm{abs}}\left(#1\right)}
\newcommand{\speedtargetirel}[2]{\speedsymbol_{#2}^{\mathrm{rel}}\left(#1\right)}
\renewcommand{\time}{t}
\newcommand{\sampletime}{\time_\mathrm{s}}
\newcommand{\thw}{\tau_\mathrm{h}}
\newcommand{\widthlanetarget}[2]{w_{#2}\left(#1\right)}

\newcommand{\cstarta}{\color{black}}  
\newcommand{\cenda}{\color{black}}     
\newcommand{\cstartb}{\color{black}}  
\newcommand{\cendb}{\color{black}}     
\newcommand{\cstartc}{\color{black}}  
\newcommand{\cendc}{\color{black}}     
\newcommand{\cstartd}{\color{black}}  
\newcommand{\cendd}{\color{black}}     
\newcommand{\cstarte}{\color{black}}  
\newcommand{\cende}{\color{black}}     
\newcommand{\cstartf}{\color{black}}  
\newcommand{\cendf}{\color{black}}     
\newcommand{\cstartg}{\color{black}}  
\newcommand{\cendg}{\color{black}}     

\begin{document}

\maketitle
\thispagestyle{empty}
\pagestyle{empty}

\begin{abstract}
	\cstarte Scenario-based methods for the assessment of Automated Vehicles (AVs) are widely supported by many players in the automotive field.
	Scenarios captured from real-world data can be used to define the scenarios for the assessment and to estimate their relevance.
	Therefore, different techniques are proposed for capturing scenarios from real-world data.
	In this paper, we propose a new method to capture scenarios from real-world data using a two-step approach.
	The first step consists in automatically labeling the data with tags.
	Second, we mine the scenarios, represented by a combination of tags, based on the labeled tags.
	One of the benefits of our approach is that the tags can be used to identify characteristics of a scenario that are shared among different type of scenarios. 
	In this way, these characteristics need to be identified only once. 
	Furthermore, the method is not specific for one type of scenario and, therefore, it can be applied to a large variety of scenarios.
	We provide two examples to illustrate the method.
	This paper is concluded with some promising future possibilities for our approach, such as automatic generation of scenarios for the assessment of automated vehicles.
	\cende
\end{abstract}

\section{Introduction}
\label{sec:introduction}

\cstartg
The development of Automated Vehicles (AVs) has made significant progress in the last years and it is expected that AVs will soon be introduced on our roads \autocite{madni2018autonomous,bimbraw2015autonomous} and become an integral part of intelligent transportation systems \autocite{eskandarian2012introduction,chanedmiston2020itsjpo}. \cendg
\cstarta An essential aspect in the development of AVs is the assessment of quality and performance aspects of the AVs, such as safety, comfort, and efficiency \autocite{bengler2014threedecades, stellet2015taxonomy}. 
Among other methods, a scenario-based approach has been proposed \autocite{elrofai2018scenario, putz2017pegasus}. 
For scenario-based assessment, proper specification of scenarios is crucial since they are directly reflected in the test cases used for the assessment \autocite{stellet2015taxonomy}. 
One approach for specifying these test cases is to base them on captured scenarios from real-world data collected on the level of individual vehicles \autocite{elrofai2018scenario, putz2017pegasus, roesener2016scenariobased, deGelder2017assessment}. 

Different techniques for capturing scenarios and driving maneuvers have been proposed in literature. 
\textcite{kasper2012oobayesnetworks} use object-oriented Bayesian networks for the recognition of 27 type of driving maneuvers. 
\textcite{krajewski2018highD} detect lane changes using lane crossings and \textcite{schlechtriemen2015lanechange} detect lane changes using a naive Bayes classifier and a hidden Markov model. 
In \autocite{xie2017driving}, random forest classifiers are used for detecting accelerating, braking, and turning with features extracted using principal component analysis, stacked sparse auto-encoders, and statistical features.
In \autocite{cara2015carcyclist}, safety-critical car-cyclist scenarios are extracted from data collected by a vehicle using several machine-learning techniques, among which support vector machines and multiple instance learning.

In this paper, we propose a new method for mining scenarios from real-world driving data using automated tagging and searching for combination of tags. 
Our method consists of two steps. 
First, the data is automatically tagged with relevant information. For example, a tag ``lane change'' is added to a vehicle at the time this vehicle is performing a lane change. 
Second, the scenarios are mined based on the aforementioned tags. \cenda
\cstartd To do this, we represent a scenario using a combination of tags and we search for this combination of tags in the tagged data from the previous step. \cendd

\cstarta The proposed method brings several benefits. 
First, by tagging the data, characteristics that are shared among different type of scenarios need to be identified only once, whereas these characteristics would be identified multiple times if each type of scenarios would be identified completely independently. \cenda
\cstartf For example, a characteristic could be the presence of a lead vehicle, so if we independently identify two different types of scenarios that consider a lead vehicle, we would identify the lead vehicle two times. \cendf
\cstarta Second, by splitting the process in two parts, i.e., the tagging and the scenario mining, the scenario mining can be applied to different types of data (e.g., data from a vehicle \autocite{paardekooper2019dataset6000km} or a measurement unit above the road \autocite{krajewski2018highD, kovvali2007video}). 
It is also possible to have manually tagged data, e.g., see \autocite{fontana2018action}. 
Third, our approach is easily scalable because additional types of scenarios can be mined by  describing them as a combination of (sequential) tags. \cenda
\cstartf Fourth, the approach reveals promising future possibilities, such as the generation of scenarios based on the mined scenarios. \cendf
\cstartg The generated scenarios can be used to define the test cases for the assessment of intelligent vehicles \autocite{stellet2015taxonomy, elrofai2018scenario, putz2017pegasus, roesener2016scenariobased, deGelder2017assessment, zhao2018evaluation}. \cendg

\cstarta In \cref{sec:problem}, we formulate the problem of scenario mining. \Cref{sec:tagging,sec:mining} describe the two steps of our proposed method, i.e., the tagging of the data and the scenario mining based on these tags. 
We illustrate the proposed scenario mining approach with few examples in \cref{sec:case study}. \cenda
\cstartf In \cref{sec:discussion}, we discuss the approach, results, and some possible future improvements. \cendf
We end this paper with conclusions and discuss next steps in \cref{sec:conclusions}. \cenda

\section{Problem formulation}
\label{sec:problem}

To formulate the scenario mining problem, we distinguish quantitative scenarios from qualitative scenarios, using the definitions of \emph{scenario} and \emph{scenario category} of \autocite{degelder2018ontology}:

\begin{definition}[Scenario]
	\label{def:scenario}
	A scenario is a quantitative description of the relevant characteristics of the ego vehicle, its activities and/or goals, its static environment, and its dynamic environment. In addition, a scenario contains all events that are relevant to the ego vehicle.
\end{definition}

\begin{definition}[Scenario category \autocite{degelder2018ontology}]
	\label{def:scenario category}
	A scenario category is a qualitative description of the ego vehicle, its activities and/or goals, its static environment, and its dynamic environment.
\end{definition}

\cstarta
A scenario category is an abstraction of a scenario and, therefore, a scenario category comprises multiple scenarios \autocite{degelder2018ontology}.
For example, the scenario category ``cut in'' comprises all possible cut-in scenarios. 
Given such a scenario category, our goal is to find all corresponding scenarios in a given data set. 
Hence, we define the scenario mining problem as follows:
\begin{problem}[Scenario mining]
	Given a scenario category, how to find all scenarios that correspond to this scenario category in a given data set?
\end{problem}
\cenda

\section{Data tagging}
\label{sec:tagging}

\begin{table*}
	\cstarta
	\centering
	\caption{\cstarta Tags that are considered in this paper.\cenda}
	\label{tab:tags}
	\begin{tabular}{llll}
		\toprule
		Subject & Description & Section & Possible tags \\ \otoprule
		Ego vehicle & Longitudinal activity & \cref{sec:longitudinal ego} & Accelerating, decelerating, cruising \\
		Ego vehicle & Lateral activity & \cref{sec:lateral ego} & Changing lane left, changing lane right, following lane \\
		Any other vehicle & Longitudinal activity & \cref{sec:longitudinal other vehicles} & Accelerating, braking, cruising \\
		Any other vehicle & Lateral activity & \cref{sec:lateral other vehicles} & Changing lane left, changing lane right, following lane \\
		Any other vehicle & Longitudinal state & \cref{sec:longitudinal state other vehicle} & In front of ego, behind ego \\
		Any other vehicle & Lateral state & \cref{sec:lateral state other vehicle} & Left of ego, right of ego, same lane as ego, unclear \\
		Any other vehicle & Lead vehicle & \cref{sec:lead vehicle} & Leader, no leader \\
		Static environment & On highway & \cref{sec:static environment} & Highway, no highway \\ 
		\bottomrule
	\end{tabular}
	\cenda
\end{table*}

\cstarta
Our method of scenario mining is divided into two steps. 
The first step consists in describing the data using \emph{tags} and the second step involves extracting the scenarios based on these tags. 
In this section, we explain how the \emph{tags} are determined. 
The scenario mining based on these tags is explained in the next section.

As described in \cref{def:scenario}, events and activities are constituents of a scenario. 
Part of the tags that we consider describe activities of the vehicles.
Therefore, we will use the definition of the term \emph{activity} of \autocite{degelder2018ontology}:\cenda
\begin{definition}[Activity]
	\label{def:activity}
	An activity quantitatively describes the time evolution of one or more state variables of an actor between two events.
\end{definition}
Because an activity starts and ends with an event, for describing the activities, we will first describe how we detect the events.

\cstarta
As an illustration, \cref{tab:tags} lists the tags that are considered for the case study in this paper. 
We distinguish between tags that describe the behavior of the ego vehicle, tags that describe the behavior and the state of any other vehicle, and tags that describe the static environment. \cenda
\begin{remark}
	\cstarta When other scenario categories are considered than the ones in our case study, other tags might be necessary. The approach for mining the scenarios using the tags, however, is general and also applies when other tags are used. \cenda 
	\cstartc For example, for this paper we do not consider other road users than vehicles, but the proposed method also works if cyclists or pedestrians are considered.\cendc
\end{remark}

In the remainder of this section, we explain how the tags of \cref{tab:tags} are assigned. 
Here, we assume that the data is sampled with a sample time of $\sampletime$.

\subsection{Longitudinal activity of the ego vehicle}
\label{sec:longitudinal ego}

We distinguish between three different longitudinal activities: ``accelerating'', ``decelerating'', and ``cruising''. 
The ego vehicle is performing either one of these activities. 
\cstartc An acceleration activity starts at an acceleration event, so we will first describe how we detect an acceleration event.
\cendc

\cstarta To extract the longitudinal events, we might simply examine whether the acceleration of the vehicle is above or below a certain threshold. 
This approach, however, would be prone to sensor noise. 
That is why we use the speed difference within a certain sample window of length $\samplehorizon>0$, where $\samplehorizon$ is an integer.
Let $\speed{\sample}$ denote the speed of the ego vehicle at sample step $\sample$. 
Next, let us define the minimum and maximum speed between the current sample step $\sample$ and $\sample-\samplehorizon$:
\begin{align}
\speedmin{\sample} &\defsym \min_{\sampledummy \in \left\{ \sample-\samplehorizon, \ldots, \sample \right\} } \speed{\sampledummy}, \\
\speedmax{\sample} &\defsym \max_{\sampledummy \in \left\{ \sample-\samplehorizon, \ldots, \sample \right\} } \speed{\sampledummy}.
\end{align}\cenda
For detecting acceleration and decelerating events, the difference between the current speed and $\speedmin{\sample}$ and $\speedmax{\sample}$ are used:
\begin{align}
\speedinc{\sample} &\defsym \speed{\sample} - \speedmin{\sample},\\
\speeddec{\sample} &\defsym \speed{\sample} - \speedmax{\sample}.
\end{align}
\cstartd \Cref{fig:explain symbols} illustrates how $\speedmin{\sample}$, $\speedmax{\sample}$, $\speedinc{\sample}$, and $\speeddec{\sample}$ are calculated.
\cendd

\setlength{\figurewidth}{.9\linewidth}
\setlength{\figureheight}{.6\linewidth}
\begin{figure}
	\centering
	\begin{tikzpicture}
	
	\definecolor{color0}{rgb}{0.12156862745098,0.466666666666667,0.705882352941177}
	
	\begin{axis}[
	axis x line*=bottom,
	axis y line*=left,
	every x tick/.style={black},
	every y tick/.style={black},
	height=\figureheight,
	tick align=outside,
	tick pos=left,
	width=\figurewidth,
	x grid style={white!69.01960784313725!black},
	xlabel={Sample},
	xmin=0, xmax=16.5,
	xtick style={color=black},
	xtick={3,13},
	xticklabels={\(\displaystyle k-k_h\),\(\displaystyle k\)},
	y grid style={white!69.01960784313725!black},
	ylabel={\(\displaystyle v\)},
	ymin=0, ymax=30,
	ytick style={color=black},
	ytick={6.48176150995277,26.6384896175217},
	yticklabels={\(\displaystyle v_{\mathrm{min}}(k)\),\(\displaystyle v_{\mathrm{max}}(k)\)}
	]
	\node at (axis cs:12.4,22.404759743144)[
	anchor= east,
	text=black,
	rotate=0.0
	]{$v^-(k)$};
	\node at (axis cs:13.6,12.3263956893595)[
	anchor= west,
	text=black,
	rotate=0.0
	]{$v^+(k)$};
	\path [draw=black, fill=white]
	(axis cs:12.4,26.6384896175217)
	--(axis cs:12.9,25.1384896175217)
	--(axis cs:12.4005,26.6384896175217)
	--(axis cs:12.4005,18.1710298687663)
	--(axis cs:12.3995,18.1710298687663)
	--(axis cs:12.3995,26.6384896175217)
	--(axis cs:11.9,25.1384896175217)
	--cycle;
	\path [draw=black, fill=white]
	(axis cs:12.4,18.1710298687663)
	--(axis cs:11.9,19.6710298687663)
	--(axis cs:12.3995,18.1710298687663)
	--(axis cs:12.3995,26.6384896175217)
	--(axis cs:12.4005,26.6384896175217)
	--(axis cs:12.4005,18.1710298687663)
	--(axis cs:12.9,19.6710298687663)
	--cycle;
	\path [draw=black, fill=white]
	(axis cs:13.6,6.48176150995277)
	--(axis cs:13.1,7.98176150995277)
	--(axis cs:13.5995,6.48176150995277)
	--(axis cs:13.5995,18.1710298687663)
	--(axis cs:13.6005,18.1710298687663)
	--(axis cs:13.6005,6.48176150995277)
	--(axis cs:14.1,7.98176150995277)
	--cycle;
	\path [draw=black, fill=white]
	(axis cs:13.6,18.1710298687663)
	--(axis cs:14.1,16.6710298687663)
	--(axis cs:13.6005,18.1710298687663)
	--(axis cs:13.6005,6.48176150995277)
	--(axis cs:13.5995,6.48176150995277)
	--(axis cs:13.5995,18.1710298687663)
	--(axis cs:13.1,16.6710298687663)
	--cycle;
	\addplot [semithick, color0, mark=*, mark size=3, mark options={solid}, only marks]
	table {%
		0 12.2953146866099
		1 18.8881758666537
		2 20.0487810681893
		3 26.6384896175217
		4 25.7460113737956
		5 19.6348856424945
		6 18.4980426894777
		7 11.6900416448795
		8 9.09754897667893
		9 7.94614004414408
		10 6.48176150995277
		11 10.4865437671727
		12 10.1138044152746
		13 18.1710298687663
		14 21.8429753436936
		15 20.3492135184764
		16 21.2848438341475
	};
	\addplot [semithick, black, dashed]
	table {%
		3 0
		3 30
	};
	\addplot [semithick, black, dashed]
	table {%
		13 0
		13 30
	};
	\addplot [semithick, black, dashed]
	table {%
		0 6.48176150995277
		16.5 6.48176150995277
	};
	\addplot [semithick, black, dashed]
	table {%
		0 26.6384896175217
		16.5 26.6384896175217
	};
	\addplot [semithick, black, dashed]
	table {%
		11.4 18.1710298687663
		14.6 18.1710298687663
	};
	\end{axis}
	
	\end{tikzpicture}
	
	\caption{\cstartd An example of a speed profile and how the variables $\speedmin{\sample}$, $\speedmax{\sample}$, $\speedinc{\sample}$, and $\speeddec{\sample}$ are calculated at a certain sample step $\sample$ with $\samplehorizon=10$.\cendd}
	\label{fig:explain symbols}
\end{figure}

First, we assume that the event at the start of the data set is a cruising event at $\sample=\sampleinit$. 
Next, we go chronologically through the data set. 
An acceleration event is happening at sample $\sample$ if any of the following conditions is true:
\begin{itemize}
	\item The vehicle is not performing an acceleration activity, i.e., the last event is not an acceleration event.
	\item \cstarta There has been a substantial speed increase between sample step $\sample-\samplehorizon$ and $\sample$, i.e., 
	\begin{equation}
	\label{eq:start acceleration}
	\speedinc{\sample} \geq \accelerationcruise\samplehorizon\sampletime,
	\end{equation} \cenda
	where $\accelerationcruise>0$ is a parameter \cstartf that describes the maximum average acceleration within the time window $\samplehorizon\sampletime$ for a cruising activity. \cendf
	\item There is no lower speed in the near future, i.e.,
	\begin{equation}
	\label{eq:acceleration condition}
	\speedmin{\sample+\samplehorizon}=\speed{\sample}.
	\end{equation}
	\item There is a substantial speed difference during the activity, i.e., 
	\begin{equation}
	\label{eq:end acceleration condition}
	\left|\speed{\sampleendinc{\sample}} - \speed{\sample}\right| > \speeddiff,
	\end{equation}
	where \cstartc $\speeddiff>0$ is the minimum speed increase and \cendc $\sampleendinc{\sample}$, \cstartf i.e., the last sample of the acceleration activity, \cendf is controlled by the parameter $\accelerationcruise$ and \cstarta equals the first sample at which the speed increase is below a threshold\cenda:
	\begin{equation}
	\label{eq:end acceleration}
	\sampleendinc{\sample} \defsym \arg \min_{\sampledummy > \sample} \left\{ \sampledummy: \speedinc{\sampledummy+\samplehorizon} < \accelerationcruise\samplehorizon\sampletime \right\}.
	\end{equation}
\end{itemize}

\cstartd
A deceleration event is detected in a similar manner as an acceleration event. 
\cstartf Now that we know the start and the end of the acceleration and deceleration activities, we simply label the remaining samples as ``cruising''. \cendf

\begin{figure}
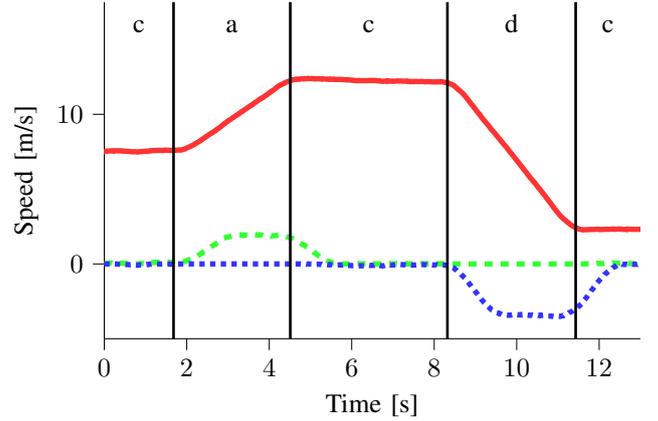

	\centering

	
	\caption{\cstarta Hypothetical speed profile and the corresponding activities cruising (c), accelerating (a), and decelerating (d). The black vertical lines represent the events times. The red solid line indicates the speed $\speed{\sample}$. The green dashed line and blue dotted line represent $\speedinc{\sample}$ and $\speeddec{\sample}$, respectively.\cenda}
	\label{fig:longitudinal activities}
\end{figure}

\cstarta%
\Cref{fig:longitudinal activities} illustrates the longitudinal activities given a hypothetical speed profile (solid red line). The algorithm above described produces the activities ``cruising'', ``accelerating'', ``cruising'', ``decelerating'', and ``cruising''.
\cenda

\setlength{\figurewidth}{\linewidth}
\setlength{\figureheight}{0.7\linewidth}

A result of the activity detection could be very short cruising activities, especially when the acceleration is around $\accelerationcruise$ or $-\accelerationcruise$. 
Therefore, all cruising activities shorter than $\samplescruising$ sample steps are removed as well as the two events that define the start and the end of the cruising activity. 
Here, we consider three possibilities:
\begin{enumerate}
	\item Before and after the cruising activity, the vehicle performs the same activity. In that case, these activities are merged.
	\item The vehicle decelerates before the cruising activity and accelerates afterwards. In that case, an acceleration event is defined at the lowest speed of the vehicle within the original cruising activity.
	\item The vehicle accelerates before the cruising activity and decelerates afterwards. In that case, a deceleration event is defined at the highest speed of the vehicle within the original cruising activity.
\end{enumerate}

\subsection{Lateral activity of the ego vehicle}
\label{sec:lateral ego}

We distinguish between three different lateral activities: ``following lane'', ``changing lane left'', and ``changing lane right''. 
To detect the lane changes, the lateral distances toward the left and right lane lines are used. 
These distances are estimated from camera images. 
The estimation is outside the scope of this paper. 
\cstartf We refer the interested reader to \cite{elfring2016effective}. \cendf
Let $\lineleft{\sample}$ and $\lineright{\sample}$ denote the distance toward the left and right lane line, respectively. 
\cstarta We use the ISO coordinate system\cenda\cstartc\footnote{\cstartc In the ISO coordinate system, the $x$-axis points to the front of the vehicle and the $y$-axis points to the left of the vehicle. \cendc\cstartd The origin of the coordinate system is often at the ground below the midpoint of the rear axle.\cendd} \autocite{iso8855}\cendc\cstarta, so $\lineleft{\sample} \geq 0$ and $\lineright{\sample} \leq 0$. 
At the moment the vehicle crosses the line, the distances to the lines will change substantially. 
For example, during a lane change to the left, the left lane line becomes the right lane line. 
Hence, we detect a left lane change \cstartd if the change in the lane line distances is more than the threshold $\lanechangethreshold>0$\cendd:
\begin{align}
\lineleft{\sample} - \lineleft{\sample-1} &> \lanechangethreshold, \label{eq:potential left a} \\
\lineright{\sample} - \lineright{\sample-1} &> \lanechangethreshold. \label{eq:potential left b}
\end{align}
Similarly, a right lane change is detected when the following conditions are satisfied:
\begin{align}
\lineleft{\sample} - \lineleft{\sample-1} &< -\lanechangethreshold, \label{eq:potential right a}\\
\lineright{\sample} - \lineright{\sample-1} &< -\lanechangethreshold. \label{eq:potential right b}
\end{align}

Once a lane change is detected using \cref{eq:potential left a,eq:potential left b,eq:potential right a,eq:potential right b}, the moment at which the lane change starts is determined. To do this, we make use of $\lineleftinc{\sample}$ and $\lineleftdec{\sample}$, which are similarly defined as $\speedinc{\sample}$ and $\speeddec{\sample}$, i.e.,
\begin{align}
\lineleftinc{\sample} \defsym \lineleft{\sample} - \min_{\sampledummy \in \left\{\sample-\samplehorizon, \ldots, \sample\right\}} \lineleft{\sampledummy}, \label{eq:line left inc}\\
\lineleftdec{\sample} \defsym \lineleft{\sample} - \max_{\sampledummy \in \left\{\sample-\samplehorizon, \ldots, \sample\right\}} \lineleft{\sampledummy}. \label{eq:line left dec}
\end{align}
Similarly, $\linerightinc{\sample}$ and $\linerightdec{\sample}$ are defined. 
If a lane change is detected at sample step $\sample$ using \cref{eq:potential left a,eq:potential left b} or \cref{eq:potential right a,eq:potential right b}, the start of this lane change is estimated in a similar manner as the start of an acceleration or deceleration activity, see \cref{eq:start acceleration}. 
The start of the lane change is at the last sample step before sample step $\sample$ at which there was not a change in either of the line distances larger than a threshold controlled by the parameter $\lanechangespeed$, i.e., for a right lane change detected at sample step $\sample$, the start is at:
\begin{equation} 
\label{eq:start lane change right}
\arg \max_{\sampledummy < \sample} \left\{ \sampledummy\colon \lineleftinc{\sampledummy} < \lanechangespeed\samplehorizon\sampletime \lor \linerightinc{\sampledummy} < \lanechangespeed\samplehorizon\sampletime \right\},
\end{equation}
where $\lor$ indicates that either one of the two conditions needs to be satisfied.

The end of a lane change is at the sample step at which either of the lane line distances increase or decrease is below a threshold. For a right lane change, this is at
\begin{multline}
\label{eq:end lane change right}
\arg \min_{\sampledummy > \sample} \left\{ \sampledummy\colon \lineleftinc{\sampledummy+\samplehorizon}<\lanechangespeed\samplehorizon\sampletime \lor \right.\\
\left. \linerightinc{\sampledummy+\samplehorizon}< \lanechangespeed\samplehorizon\sampletime \right\}
\end{multline}
For a left lane change, the start and end is defined by substituting $-\lineleftdec{\cdot}$ and $-\linerightdec{\cdot}$ for $\lineleftinc{\cdot}$ and $\linerightinc{\cdot}$, respectively, in \cref{eq:start lane change right,eq:end lane change right}.
\cenda

\setlength{\figurewidth}{\linewidth}
\setlength{\figureheight}{0.6\linewidth}
\begin{figure}
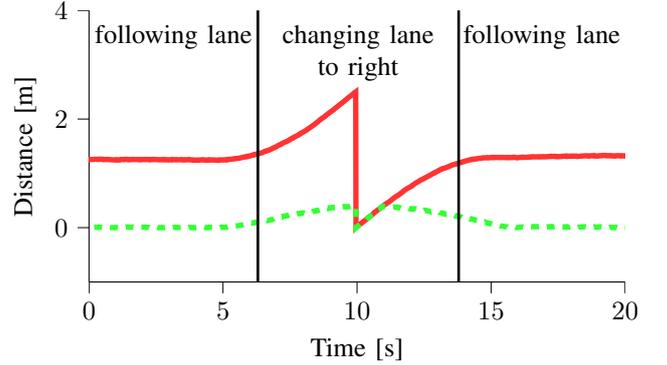

	\centering

	
	\caption{\cstarta The red line represents the hypothetical distance toward the left lane line $\lineleft{\sample}$ during a right lane change of the ego vehicle. The black vertical lines indicate the time instants of the events at the start and the end of the lane change. The green dashed line line represents $\lineleftinc{\sample}$.\cenda}
	\label{fig:ego lane change}
\end{figure}

\cstarta
\Cref{fig:ego lane change} illustrates a hypothetical lane change of the ego vehicle. 
It shows that the distance towards the left lane line changes when the ego vehicle crosses the line, such that the conditions \cref{eq:potential right a,eq:potential right b} are satisfied. In \cref{fig:ego lane change}, events at the start and the end of the lane change are denoted by the black vertical lines.

\begin{figure}
	\centering
	\includegraphics[width=\linewidth]{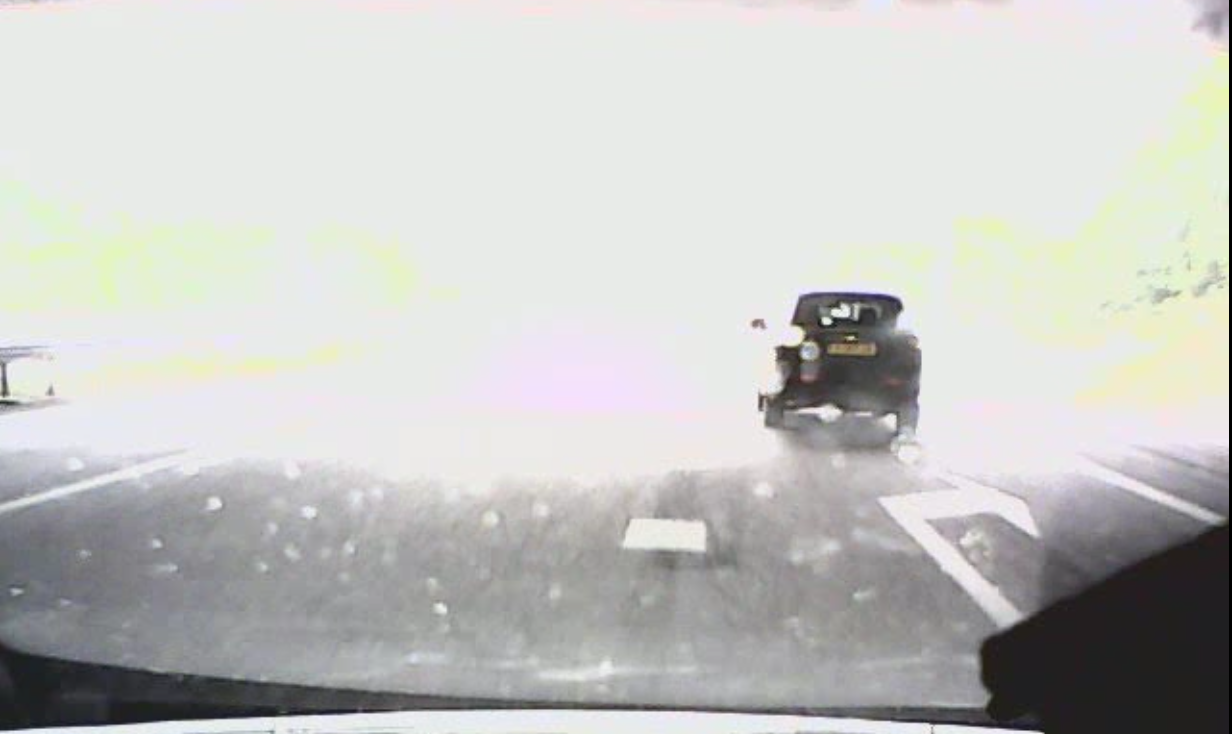}
	\caption{\cstarta The ego vehicle passes a flyover during daytime while performing a lane change. This causes glare such that the distance to the lane lines are not available. 
		\cenda}
	\label{fig:glare}
\end{figure}

\begin{remark}
	It might happen that there is no accurate measurement of the lane line distances available at a certain sample step $\sample$. For example, in \cref{fig:glare}, there is no line information while the ego vehicle performs a lane change. By using the next available line distances instead of $\lineleft{\sample}$ and $\lineright{\sample}$ and the previous available line distances instead of $\lineleft{\sample-1}$ and $\lineright{\sample-1}$ in \cref{eq:potential left a,eq:potential left b,eq:potential right a,eq:potential right b}, our algorithm is still able to detect lane changes.
\end{remark}
\cenda

\subsection{Longitudinal activity of other vehicle}
\label{sec:longitudinal other vehicles}

The longitudinal activities of other vehicles are estimated in a similar manner as for the ego vehicle. However, instead of the speed of the ego vehicle, $\speed{\sample}$, the speed of the other vehicles is used. The ego vehicle measures the relative speed of other vehicles. Let $\speedtargetirel{\sample}{\indextarget}$ denote the relative speed of the $\indextarget$-th vehicles at sample $\sample$. The absolute speed of other vehicles is estimated by adding $v(k)$ to the estimated relative speed:
\begin{equation}
\speedtargetiabs{\sample}{\indextarget} = \speedtargetirel{\sample}{\indextarget} + \speed{\sample}.
\end{equation}
To compute the longitudinal activities of the $\indextarget$-th vehicle, the approach outlined in \cref{sec:longitudinal ego} is used with $\speedtargetiabs{\sample}{\indextarget}$ substituted for $\speed{\sample}$. 

\begin{remark}
	\label{rem:no target}
	\cstarta Typically, $\speedtargetirel{\sample}{\indextarget}$ is obtained by fusing the outputs of several sensors \autocite{elfring2016effective}. 
	If $\speedtargetirel{\sample}{\indextarget}$ is not available, e.g., because the vehicle moved out of the view of the ego vehicle's sensors, there are no activities estimated for the $\indextarget$-th vehicle at sample step $\sample$. \cenda
	\cstartc Consequently, no tags are applied for the $\indextarget$-th vehicle at sample step $\sample$. This applies for all tags of the other vehicles that are mentioned in \cref{tab:tags}.\cendc
\end{remark}

\subsection{Lateral activity of other vehicle}
\label{sec:lateral other vehicles}

\cstartc
For the lane changes of other vehicles, only the lane changes to and from the ego vehicle's lane are considered.
To detect a lane change of the $\indextarget$-th vehicle, we use the distance of the $\indextarget$-th vehicle toward the ego vehicle's left and right lane lines, denoted by $\linelefttarget{\sample}{\indextarget}$ and $\linerighttarget{\sample}{\indextarget}$, respectively. \cendc
\cstartd $\linelefttarget{\sample}{\indextarget}$ and $\linerighttarget{\sample}{\indextarget}$ are determined by subtracting the estimated lane line positions from the estimated lateral position of the $\indextarget$-th vehicle. 
The lane line positions are based on the estimated shape of the lane lines. 
For more details, we refer the reader to \autocite{elfring2016effective}. \cendd
\cstartc We define $\linelefttargetinc{\sample}{\indextarget}$, $\linerighttargetinc{\sample}{\indextarget}$, $\linelefttargetdec{\sample}{\indextarget}$, and $\linerighttargetdec{\sample}{\indextarget}$ similar as $\lineleftinc{\sample}$ and $\lineleftdec{\sample}$ in \cref{eq:line left inc,eq:line left dec}.

A lane change is detected if the vehicle crosses either of the two lane lines.
There are four possible ways this can happen.
For now, we consider a right lane change toward the ego vehicle's lane.
A right lane change of the $\indextarget$-th vehicle toward the ego vehicle's lane is detected at sample step $\sample$ if the vehicle is not already changing lane and
\begin{equation}
\label{eq:detect lane change target}
\linelefttarget{\sample-1}{\indextarget} \leq 0 \land \linelefttarget{\sample}{\indextarget} > 0,
\end{equation}
where $\land$ indicates that both of the two conditions need to be satisfied.

To determine the start of the lane change, the lateral speed should be below the threshold $\lanechangespeed$ or --- in case the vehicle changes several lanes --- the lateral movement should be above a certain threshold (controlled by $\factorgoalmax$).
Because it might happen that the lateral speed is below the threshold during the whole lane change, a minimum lateral movement is considered as well (controlled by $\factorgoalmin$). 
As a result, the start of a right lane change toward the ego vehicle's lane is estimated to occur at sample step
\begin{multline}
\label{eq:start lane change target}
\arg \max_{\sampledummy < \sample} \left\{ \sampledummy: \linelefttarget{\sampledummy}{\indextarget} < -\factorgoalmax\widthlanetarget{\sample}{\indextarget} \lor \right. \\
\left. \left( \linelefttargetinc{\sampledummy}{\indextarget} < \lanechangespeed\samplehorizon\sampletime \land \linelefttarget{\sampledummy}{\indextarget} < -\factorgoalmin\widthlanetarget{\sample}{\indextarget} \right) \right\}.
\end{multline}
Here, $\widthlanetarget{\sample}{\indextarget}=\linelefttarget{\sample}{\indextarget}-\linerighttarget{\sample}{\indextarget}$ is the estimated lane width. 
The end of the same lane change is estimated, in a similar way, to occur at sample step:
\begin{multline}
\label{eq:end lane change target}
\arg \max_{\sampledummy > \sample} \left\{ \sampledummy: \linelefttarget{\sampledummy}{\indextarget} > \factorgoalmax\widthlanetarget{\sample}{\indextarget} \lor \right. \\
\left. \left(\linelefttargetinc{\sampledummy+\samplehorizon}{\indextarget} < \lanechangespeed\samplehorizon\sampletime \land \linelefttarget{\sampledummy}{\indextarget} > \factorgoalmin\widthlanetarget{\sample}{\indextarget}\right) \right\}.
\end{multline}

\cendc
\cstartf A right lane change from the ego vehicle's lane and a left lane change from or to the ego vehicle's lane are determined in a similar manner. \cendf

\cstartc
\subsection{Longitudinal state of other vehicle}
\label{sec:longitudinal state other vehicle}

For the longitudinal state of any other vehicle, two possibilities are considered: in front of the ego vehicle or behind the ego vehicle. 
Let the longitudinal position at sample step $\sample$ of the $\indextarget$-th vehicle relative to the ego vehicle be denoted by $\lontarget{\sample}{\indextarget}$. 
The tag ``in front of ego'' applies when $\lontarget{\sample}{\indextarget}> 0$ and the tag ``behind ego'' applies when $\lontarget{\sample}{\indextarget}\leq 0$.

\subsection{Lateral state of other vehicle}
\label{sec:lateral state other vehicle}

Four different possibilities are considered for the lateral state of any other vehicle. 
The lateral state is based on the estimated distance of the other vehicle toward the ego vehicle's lane lines, see \cref{tab:lateral state other vehicle}. 
The situation of $\linelefttarget{\sample}{\indextarget} < 0$ and $\linerighttarget{\sample}{\indextarget} \geq 0$ would mean that the other vehicle is left of the left lane line and right of the right lane line, so it is unclear in which lane the vehicle is.

\begin{table}
	\centering
	\caption{\cstartc Lateral state based on $\linelefttarget{\sample}{\indextarget}$ and $\linerighttarget{\sample}{\indextarget}$.\cendc}
	\label{tab:lateral state other vehicle}
	\cstartc
	\begin{tabular}{lcc}
		\toprule
		& $\linelefttarget{\sample}{\indextarget} < 0$ & $\linelefttarget{\sample}{\indextarget} \geq 0$ \\ \otoprule		$\linerighttarget{\sample}{\indextarget} < 0$ & Left of ego & Same lane as ego \\
		$\linerighttarget{\sample}{\indextarget} \geq 0$ & Unclear & Right of ego \\
		\bottomrule
	\end{tabular}
	\cendc
\end{table}

\subsection{Lead vehicle}
\label{sec:lead vehicle}

Two possibilities are considered: a vehicle is a lead vehicle (the tag ``leader'' applies) or not (the tag ``no leader'' applies). A vehicle $\indextarget$ is considered as a lead vehicle at sample step $\sample$ if all of the following conditions are satisfied:
\begin{itemize}
	\item The vehicle is in front of the ego vehicle, i.e., $\lontarget{\sample}{\indextarget}>0$.
	\item The vehicle drives in the same lane as the ego vehicle, i.e., $\linelefttarget{\sample}{\indextarget} \geq 0$ and $\linerighttarget{\sample}{\indextarget} < 0$.
	\item The time headway of the ego vehicle toward the other vehicle, i.e., $\lontarget{\sample}{\indextarget}/\speed{\sample}$ is less than the parameter $\thw>0$.
	\item There is no other vehicle that is closer to the ego vehicle while satisfying the above conditions, i.e., $\lontarget{\sample}{\indextarget} \leq \lontarget{\sample}{\indextargetother}$ for all $\indextargetother$-th vehicles that satisfy the above conditions.
\end{itemize}
\cendc

\subsection{Static environment}
\label{sec:static environment}

\cstartc
The aspect of the static environment that is considered in this paper is whether the ego vehicle drives on the highway or not. The location of the ego vehicle, based on GPS measurements, is used to determine the road the ego vehicle is driving on based on OpenStreetMaps\footnote{\cstartc See \url{https://www.openstreetmap.org/}.\cendc}. If the road is classified as ``motorway'' (see \autocite{osm_highway} for all possibilities), the tag ``highway'' is applied. Otherwise, the tag ``no highway'' is used.
\cendc

\section{Mining scenarios using tags}
\label{sec:mining}

\cstartd
For the scenario mining, we formulate a scenario category using a combination of tags.
As an example, \cref{fig:cutin formulation tags} shows how the scenario category ``cut in'' can be formulated using tags.
To further structure the tags, we formulate a scenario category as a sequence of \emph{items} where each \emph{item} corresponds to a combination of tags for all relevant subjects.
The number of items may vary from scenario category to scenario category.
The scenario category ``cut in'' in \cref{fig:cutin formulation tags} contains two items and considers a vehicle other than the ego vehicle that changes lane (other vehicle, item 1 and 2) and becomes the lead vehicle (other vehicle, item 2).
In the meantime, the ego vehicle follows its lane (ego vehicle, items 1 and 2) and the scenario category only considers highway driving (static environment, items 1 and 2).
When describing the tags for each item, logical AND, OR, or NOT rules may be used. 
For example, for the other vehicle in \cref{fig:cutin formulation tags}, either the tag ``changing lane left'' \emph{or} the tag ``changing lane right'' needs to apply.
\cendd

\begin{figure}
	\centering
	\setlength{\itemwidth}{5.5em}
	\begin{tikzpicture}
	\node[block, text width=\subjectwidth-1em, minimum width=\subjectwidth, fill=egocolor] at (-\descriptionwidth, 0) {Ego vehicle};
	\node[block, text width=\descriptionwidth-1em, minimum width=\descriptionwidth, fill=egocolor] at (0, 0) {Lateral activity};
	\node[tagitemtwo, fill=egocolor] at (0, 0) {Following lane};
	
	\node[block, text width=\subjectwidth-1em, minimum width=\subjectwidth, fill=othervehicle, minimum height=2\tagheight] at (-\descriptionwidth, -\tagheight-\tagsep) {Other vehicle};
	\node[block, text width=\descriptionwidth-1em, minimum width=\descriptionwidth, fill=othervehicle] at (0, -\tagheight-\tagsep) {Lateral activity};
	\node[tagitemtwo, fill=othervehicle] at (0, -\tagheight-\tagsep) {Changing lane left OR Changing lane right};
	
	\node[block, text width=\descriptionwidth-1em, minimum width=\descriptionwidth, fill=othervehicle] at (0, -2\tagheight-\tagsep) {Lead vehicle};
	\node[tagitem, fill=othervehicle] at (0, -2\tagheight-\tagsep) {No leader};
	\node[tagitem, fill=othervehicle] at (\itemwidth, -2\tagheight-\tagsep) {Leader};
	
	\node[block, text width=\subjectwidth-1em, minimum width=\subjectwidth, fill=staticenvironment] at (-\descriptionwidth, -3\tagheight-2\tagsep) {Static environment};
	\node[block, text width=\descriptionwidth-1em, minimum width=\descriptionwidth, fill=staticenvironment] at (0, -3\tagheight-2\tagsep) {On highway};
	\node[tagitemtwo, fill=staticenvironment] at (0, -3\tagheight-2\tagsep) {Highway};
	
	\foreach \i in {1, 2} {%
		\node[minimum width=\itemwidth, align=center, minimum height=\itempos, anchor=north east] at (\i\itemwidth, \itempos) {Item \i};
		\draw[showitem] (\i\itemwidth, 0) -- (\i\itemwidth, \itempos);
	}
	\draw[showitem] (0, 0) -- (0, \itempos);
	\end{tikzpicture}
	\caption{\cstartd Formulation of the scenario category ``cut in'' using tags.\cendd}
	\label{fig:cutin formulation tags}
\end{figure}

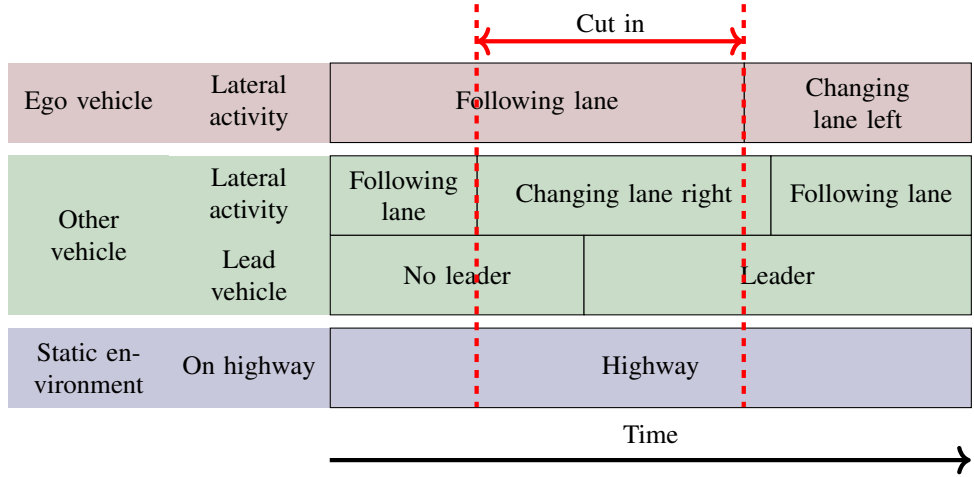
\begin{figure*}
	\centering
	\setlength{\tagtotalwidth}{24em}
	\setlength{\egovehiclefollowing}{15.5em}
	\begin{tikzpicture}
	\node[block, text width=\subjectwidth-1em, minimum width=\subjectwidth, fill=egocolor] at (-\descriptionwidth, 0) {Ego vehicle};
	\node[block, text width=\descriptionwidth-1em, minimum width=\descriptionwidth, fill=egocolor] at (0, 0) {Lateral activity};
	\node[tag, minimum width=\egovehiclefollowing, fill=egocolor] at (0, 0) {Following lane};
	\node[tag, minimum width=\tagtotalwidth-\egovehiclefollowing, text width=\tagtotalwidth-\egovehiclefollowing-1em, fill=egocolor] at (\egovehiclefollowing, 0) {Changing lane left};
	
	\node[block, text width=\subjectwidth-1em, minimum width=\subjectwidth, fill=othervehicle, minimum height=2\tagheight] at (-\descriptionwidth, -\tagheight-\tagsep) {Other vehicle};
	\node[block, text width=\descriptionwidth-1em, minimum width=\descriptionwidth, fill=othervehicle] at (0, -\tagheight-\tagsep) {Lateral activity};
	\node[tag, text width=\otherfollowing-1em, minimum width=\otherfollowing, fill=othervehicle] at (0, -\tagheight-\tagsep) {Following lane};
	\node[tag, text width=\otherchanging-1em, minimum width=\otherchanging, fill=othervehicle] at (\otherfollowing, -\tagheight-\tagsep) {Changing lane right};
	\node[tag, text width=\tagtotalwidth-\otherfollowing-\otherchanging-1em, minimum width=\tagtotalwidth-\otherfollowing-\otherchanging, fill=othervehicle] at (\otherfollowing+\otherchanging, -\tagheight-\tagsep) {Following lane};
	
	\node[block, text width=\descriptionwidth-1em, minimum width=\descriptionwidth, fill=othervehicle] at (0, -2\tagheight-\tagsep) {Lead vehicle};
	\node[tag, minimum width=\othernolead, fill=othervehicle] at (0, -2\tagheight-\tagsep) {No leader};
	\node[tag, minimum width=\tagtotalwidth-\othernolead, fill=othervehicle] at (\othernolead, -2\tagheight-\tagsep) {Leader};
	
	\node[block, text width=\subjectwidth-1em, minimum width=\subjectwidth, fill=staticenvironment] at (-\descriptionwidth, -3\tagheight-2\tagsep) {Static environment};
	\node[block, text width=\descriptionwidth-1em, minimum width=\descriptionwidth, fill=staticenvironment] at (0, -3\tagheight-2\tagsep) {On highway};
	\node[tag, minimum width=\tagtotalwidth, fill=staticenvironment] at (0, -3\tagheight-2\tagsep) {Highway};
	
	\draw[cutinline, dashed] (\otherfollowing, \cutinheight) -- (\otherfollowing, -4\tagheight-2\tagsep);
	\draw[cutinline, dashed] (\egovehiclefollowing, \cutinheight) -- (\egovehiclefollowing, -4\tagheight-2\tagsep);
	\draw[cutinline, <->] (\otherfollowing, \arrowheight) -- (\egovehiclefollowing, \arrowheight);
	\node[anchor=north west, text width=\egovehiclefollowing-\otherfollowing-1em, minimum width=\egovehiclefollowing-\otherfollowing, minimum height=\cutinheight-\arrowheight, align=center] at (\otherfollowing, \cutinheight) {Cut in};
	
	\draw[timeline, ->] (0, -4\tagheight-2\tagsep-\timepos) -> (\tagtotalwidth, -4\tagheight-2\tagsep-\timepos);
	\node[minimum width=\tagtotalwidth, align=center, anchor=north west, minimum height=\timepos] at (0, -4\tagheight-2\tagsep) {Time};
	\end{tikzpicture}
	
	\caption{\cstartc Example of tags describing a cut in. Note that only the tags that are relevant for the cut in, as defined in \cref{fig:cutin formulation tags}, are shown. \cendc\cstartf Furthermore, whereas there are multiple other vehicles around the ego vehicle, only the other vehicle that performs the cut in is shown.\cendf}
	\label{fig:tags cut in}
\end{figure*}

\cstartb
The scenarios are mined by searching for matches of the defined items within the tags of the data set. 
This searching is subject to two rules:
\begin{enumerate}
	\item For each item, there needs to be a match for all relevant subjects \emph{at the same sample time}.
	\item The different items need to occur \emph{right after each other}. 
\end{enumerate}\cendb
\cstartd To continue the example of the scenario category ``cut in'', \cref{fig:tags cut in} shows a part of labeled data in which a cut-in scenario is found. \cendd
\cstartb The two vertical dashed lines indicate the start and the end of the cut in that is defined in \cref{fig:cutin formulation tags}.
\cendb

\section{Case study}
\label{sec:case study}

\begin{figure}
	\centering
	\includegraphics[width=.8\linewidth]{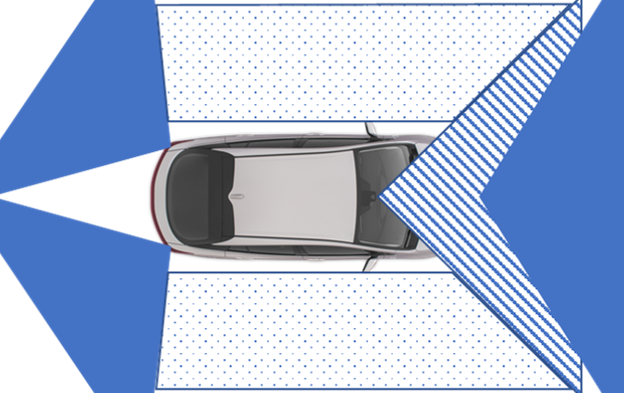}
	\caption{\cstartd Schematic representation of the field of view of the three radars (solid area) and the camera (area filled with lines) that the ego vehicle is equipped with. The positions of vehicles on the left or the right of the ego vehicle (dotted area) are predicted based on previous measurements. \cendd}
	\label{fig:sensors}
\end{figure}

Here we illustrate the proposed method by applying it to the data set described in \autocite{paardekooper2019dataset6000km}.
\cstartf The data have been recorded from a single vehicle in which different drivers were asked to drive a prescribed route.
The majority of the route is on the highway. \cendf
\cstartd To measure the surrounding traffic, the vehicle is equipped with three radars and one camera, as shown in \cref{fig:sensors}.
The images from the camera are used to estimate the lane line distances \autocite{elfring2016effective}.
Furthermore, the surrounding traffic is measured by fusing the data of the radars and the camera \autocite{elfring2016effective}.
While fusing the data of the different sensors, the position of the vehicles that disappear from the sensors' field of view on the left and right of the ego vehicle, see the dotted areas in \cref{fig:sensors}, are predicted until the vehicles appear again in the sensors' field of view.
In total, four hours of driving are analyzed.
\cendd

\cstartd To illustrate the proposed scenario mining approach, two different scenario categories are considered: ``cut in'' and ``overtaking before lane change''.
\Cref{fig:cutin formulation tags,fig:overtaking template} show the formulation of these scenario categories using tags.
\Cref{tab:parameters} lists the values of the parameters that are used for the tagging of the data.
\cendd

\begin{figure*}
	\centering
	\setlength{\descriptionwidth}{7em}
	\begin{tikzpicture}
	\node[block, text width=\subjectwidth-1em, minimum width=\subjectwidth, fill=egocolor] at (-\descriptionwidth, 0) {Ego vehicle};
	\node[block, text width=\descriptionwidth-1em, minimum width=\descriptionwidth, fill=egocolor] at (0, 0) {Lateral activity};
	\node[tagitemtwo, fill=egocolor] at (0, 0) {Following lane};
	\node[tagitemtwo, fill=egocolor] at (2\itemwidth, 0) {Changing lane left};
	
	\node[block, text width=\subjectwidth-1em, minimum width=\subjectwidth, fill=othervehicle, minimum height=2\tagheight] at (-\descriptionwidth, -\tagheight-\tagsep) {Other vehicle};
	\node[block, text width=\descriptionwidth-1em, minimum width=\descriptionwidth, fill=othervehicle] at (0, -\tagheight-\tagsep) {Lateral state};
	\node[tagitemthree, fill=othervehicle] at (0, -\tagheight-\tagsep) {Left of ego};
	\node[tagitem, fill=othervehicle] at (3\itemwidth, -\tagheight-\tagsep) {Same lane as ego};
	
	\node[block, text width=\descriptionwidth-1em, minimum width=\descriptionwidth, fill=othervehicle] at (0, -2\tagheight-\tagsep) {Longitudinal state};
	\node[tagitem, fill=othervehicle] at (0, -2\tagheight-\tagsep) {Behind ego};
	\node[tagitemthree, fill=othervehicle] at (\itemwidth, -2\tagheight-\tagsep) {In front of ego};
	
	\node[block, text width=\subjectwidth-1em, minimum width=\subjectwidth, fill=staticenvironment] at (-\descriptionwidth, -3\tagheight-2\tagsep) {Static environment};
	\node[block, text width=\descriptionwidth-1em, minimum width=\descriptionwidth, fill=staticenvironment] at (0, -3\tagheight-2\tagsep) {On highway};
	\node[tagitemfour, fill=staticenvironment] at (0, -3\tagheight-2\tagsep) {Highway};
	
	\foreach \i in {1, 2, 3, 4} {%
		\node[minimum width=\itemwidth, align=center, minimum height=\itempos, anchor=north east] at (\i\itemwidth, \itempos) {Item \i};
		\draw[showitem] (\i\itemwidth, 0) -- (\i\itemwidth, \itempos);
	}
	\draw[showitem] (0, 0) -- (0, \itempos);
	\end{tikzpicture}
	\caption{\cstartd Formulation of the scenario category ``overtaking before lane change'' using tags.\cendd}
	\label{fig:overtaking template}
\end{figure*}

\begin{table}
	\centering
	\caption{\cstartc Values of parameters used in the case study. \cendc}
	\label{tab:parameters}
	\cstartc
	\begin{tabularx}{\linewidth}{lXl}
		\toprule
		Parameter & Description & Value \\ \otoprule
		$\sampletime$ & Sample time & \SI{0.01}{\second} \\
		$\samplehorizon$ & Sample window & 100 \\
		$\accelerationcruise$ & Threshold determining the start and end of an acceleration or deceleration activity & \SI{0.1}{\meter\per\second\squared} \\
		$\speeddiff$ & Minimum speed increase/decrease for an acceleration/deceleration activity & \SI{1}{\meter\per\second} \\
		$\samplescruising$ & Minimum number of samples for cruising activity & 400 \\
		$\lanechangethreshold$ & A lane change is detected when the difference between consecutive lane line distances is larger than this threshold & \SI{1}{\meter} \\
		$\lanechangespeed$ & Threshold determining the start and end of a lane change & \SI{0.25}{\meter\per\second} \\
		$\factorgoalmax$ & Maximum factor of the lane width for a lane change of any other vehicle & 0.5 \\
		$\factorgoalmin$ & Minimum factor of the lane width for a lane change of any other vehicle & 0.1 \\
		\bottomrule
	\end{tabularx}
	\cendc
\end{table}

\cstartd The results of the scenario mining are presented in \cref{tab:results}.
A false negative (FN) means that a scenario that occurred is not detected and a false positive (FP) means that the scenario mining detects a scenario whereas this scenario does not occur.
The true positives (TP) are the scenarios that are correctly detected.
The recall is the ratio of the number of true positives (TP) and the total number of scenarios that occur (TP+FN) and the precision is the ratio of the number of true positives (TP) and the total number of detected scenarios (FP+TP).
The F1 score is the harmonic mean of the recall and the precision: \cendd
\cstartf \begin{equation}
\fonescore = 2 \cdot \frac{\precision\cdot\recall}{\precision+\recall}.
\end{equation} \cendf

\begin{table}
	\centering
	\caption{\cstartd Results of the scenario mining.\cendd}
	\label{tab:results}
	\cstartd
	\begin{tabularx}{\linewidth}{Xr<{\hspace{-1pt}}r<{\hspace{-1pt}}r<{\hspace{-1pt}}r<{\hspace{-1pt}}r<{\hspace{-1pt}}r<{\hspace{-1pt}}}
		\toprule
		Scenario category & FN & FP & TP & Recall & Precision & F1 score \\ \otoprule
		Cut in & 3 & 3 & 33 & \SI{92}{\percent} & \SI{92}{\percent} & \SI{92}{\percent}  \\
		Overtaking before lane change & 1 & 0 & 18 & \SI{95}{\percent} & \SI{100}{\percent} & \SI{97}{\percent} \\
		\bottomrule
	\end{tabularx}
	\cendd
\end{table}

\cstartd As listed in \cref{tab:results}, 33 out of 36 cut ins are correctly detected and 3 out of the 36 detected cut ins are incorrect. 
This results in an F1 score of \SI{92}{\percent}.
For the scenario category ``overtaking before lane change'', 18 out of 19 scenarios are correctly detected and there are no scenarios incorrectly detected.
This results in an F1 score of \SI{97}{\percent}. \cendd

\cstartf\section{Discussion}\cendf
\label{sec:discussion}

\cstartd The false detections are a result of inaccurate or missing data.
For example, in case of the four false negatives, the other vehicle is not detected at the time of the cut in or overtaking.
For one cut in, this is because another vehicle obstructs the view toward the vehicle at the moment of the cut in. 
For the other three false negatives, the other vehicles appear from the sensor's blind spot (dotted area in \cref{fig:sensors}).
The three false positives of the cut-in scenario are a result of inaccurate measurements of the lane line distances. \cendd
\cstartf On the one hand, it might be interpreted as that the false detections are due to limitations of the data.
On the other hand, for future work, we can expand our work to deal with these limitations of the data.
For example, using techniques used for correcting the interpretation of natural language \autocite{hull1982experiments}, we might be able to correct wrong tags or to add missing tags. \cendf

\cstartf To mine scenarios from a scenario category, the scenario category needs to be represented by a a certain combination of tags, such as shown in \cref{fig:cutin formulation tags,fig:overtaking template}.
Provided that there are no new tags required, there are no new algorithms required for mining scenarios from new scenario categories. 
As a result, it is relatively straightforward to apply the proposed approach for mining scenarios from other scenario categories than the ones presented in our case study. \cendf
\cstarte Future work includes more tags, e.g., ``turning left'' or ``turning right'', and to consider more actors, e.g., pedestrians and cyclists. This will enable the mining of many more scenarios. \cende

\cstartf For future research, the analogy between the proposed scenario mining and natural language processing (NLP) could be explored. 
In NLP, natural language is analyzed by searching for certain combination of words or syllables. 
Similarly, we are searching for certain combinations of tags. \cendf
\cstarte In NLP, n-gram models are successfully used to correct \autocite{hull1982experiments} and predict \autocite{brown1992class} words and to generate text \autocite{oh2002stochastic}; so n-gram models might be used to correct and predict tags and to generate new scenarios for the assessment of automated vehicles. \cende

\section{Conclusions}
\label{sec:conclusions}

\cstarte
For the scenario-based assessment of automated vehicles, scenarios captured from real-world data collected on the level of individual vehicles can be used to define the tests.
We have proposed a two-step approach for mining real-world scenarios from a data set.
The first step consists in labeling the data with tags that describe, e.g., the lateral and longitudinal activities of the different actors.
The second step mines the scenarios by searching for particular combinations of tags.
We have illustrated the approach with two examples, a cut in and an overtaking before a lane change. \cende
\cstartf These examples demonstrated that the proposed approach is suitable for mining scenarios from real-world data.
Future work includes labeling the data with more tags and exploring the possibilities of using techniques that are used in the field of natural language processing.
\cendf


\bibliographystyle{ieeetran}
\bibliography{references}

\end{document}